\RequirePackage{fix-cm}
\documentclass[conference]{IEEEtran}
\usepackage{graphicx}
\usepackage{cite}
\usepackage{soul}
\usepackage{amsmath,amssymb,amsfonts}
\usepackage{algorithmic}
\usepackage{textcomp}
\usepackage{xcolor}
\usepackage{listings}
\usepackage{moresize}
\usepackage{subcaption}
\usepackage{float}
\usepackage{caption}
\usepackage[utf8]{inputenc}
\usepackage{array}
\usepackage{hyperref}

\newcolumntype{M}[1]{>{\centering\arraybackslash}m{#1}}
\newcolumntype{P}[1]{>{\centering\arraybackslash}p{#1}}
\begin{document}

\title{Using Conditional Generative Adversarial Networks to Reduce the Effects of Latency in Robotic Telesurgery}


\author{
    
    \IEEEauthorblockN{Neil Sachdeva}
    \IEEEauthorblockA{
        Machine Perception and Cognitive Robotics Lab \\
        Florida Atlantic University\\
        Pine Crest School\\
        Boca Raton, Florida, USA \\
        neil.sachdeva@gmail.com
    }
   
    \and
    \IEEEauthorblockN{Misha Klopukh}
    \IEEEauthorblockA{
        Machine Perception and Cognitive Robotics Lab \\
        Florida Atlantic University\\
        Boca Raton, Florida, USA \\
        mishakmak@gmail.com
    }
    \and
    
    \IEEEauthorblockA{}
    \IEEEauthorblockA{}
    \and
    \IEEEauthorblockA{}
    \IEEEauthorblockA{}
    \and
    \\
    \IEEEauthorblockN{Rachel St. Clair}
    \IEEEauthorblockA{
        Machine Perception and Cognitive Robotics Lab \\
        Florida Atlantic University\\
        Boca Raton, Florida, USA \\
        rstclair2021@fau.edu
    }
    \and
    \\
    \IEEEauthorblockN{William Edward Hahn}
    \IEEEauthorblockA{
        Machine Perception and Cognitive Robotics Lab\\
        Florida Atlantic University\\
        Boca Raton, Florida, USA \\
        williamedwardhahn@gmail.com
    }
    
}
\maketitle 

\begin{abstract}
The introduction of surgical robots brought about advancements in surgical procedures. The applications of remote telesurgery range from building medical clinics in underprivileged areas, to placing robots abroad in military hot-spots where accessibility and diversity of medical experience may be limited. Poor wireless connectivity may result in a prolonged delay, referred to as latency, between a surgeon’s input and action a robot takes. In surgery, any micro-delay can injure a patient severely and in some cases, result in fatality. One way to increase safety is to mitigate the effects of latency using deep learning aided computer vision. While the current surgical robots use calibrated sensors to measure the position of the arms and tools, in this work we present a purely optical approach that provides a measurement of the tool position in relation to the patient's tissues. This research aimed to produce a neural network that allowed a robot to detect its own mechanical manipulator arms. A conditional generative adversarial network (cGAN) was trained on 1107 frames of a mock gastrointestinal robotic surgery from the 2015 EndoVis Instrument Challenge and corresponding hand-drawn labels for each frame. When run on new testing data, the network generated near-perfect labels of the input images which were visually consistent with the hand-drawn labels and was able to do this in 299 milliseconds. These accurately generated labels can then be used as simplified identifiers for the robot to track its own controlled tools. These results show potential for conditional GANs as a reaction mechanism such that the robot can detect when its arms move outside the operating area in a patient. This system allows for more accurate monitoring of the position of surgical instruments in relation to the patient's tissue, increasing safety measures that are integral to successful telesurgery systems.

\end{abstract}


\hrule

\let\thefootnote\relax\footnotetext{\textit{Full article can be found at the following link: \textbf{\href{https://link.springer.com/article/10.1007/s11701-020-01149-5}{Journal of Robotic Surgery}}\\
DOI: 10.1007/s11701-020-01149-5
}}

\section{Introduction}
Surgical robots, such as the da Vinci Surgical System allow for surgeons to perform minimally invasive surgeries with pinpoint accuracy and complete maneuverability. In a typical robotic surgery system the surgeon’s console is directly wired to the robot and a screen that shows a live feed of the robotic arms inside the patient. 

For surgical robots to have full reliability in a remote setup far from the operating surgeon, they need to be able to continue operating even in scenarios where network connection is unreliable, as any microsecond delay can potentially result in a serious accident. In addition, no networks have 100\% reliability, so there is a lag time where a video feed could freeze or a command to move the robot is not received, in which case the robot would continue moving even if a patient got in the way. These risks have discouraged the expansive use of the practice and while there are currently operating remote surgeries \cite{nature_2001}, they cannot be utilized on a large scale because of the potential dangers associated with latency \cite{bernal2017comparative, jin2019towards}. In studies that measured the effects of latency on a surgical performance \cite{perez_xu_chauhan_tanaka_simpson_abdul-muhsin_smith_2015,latency_on_surgery} it was determined that exceeding 300 ms in latency causes "measurable deterioration of performance" in surgical accuracy and thus are not practical for transcontinental surgical applications requiring efficient and reliable reaction metrics\cite{pings}. Addressing the concern of latency is the primary concern of this work for aiding telesurgery reliability and practicality in the field.

By implementing a computer vision aided system to serve as an intermediary between the robot and the surgeon, the robot is no longer solely dependent on the surgeon and thus the effects of input lag are mitigated - specifically during the time it takes for a command to reach the robot which is when the on-board autonomous system can take control. In real-world applications, a robot would be stationed in a remote location and a doctor would be at their control station located in their own office. The neural network would be loaded onto the surgical robot’s on-board computer and would be able to take control of the robot’s arms whenever necessary. If a network interruption occurred, the neural network could recognize the robotic arms moving towards a dangerous position and override the robot’s controls, forcing it to stop. This system has the potential to accurately monitor the surgical instruments in relation to the patient's tissue. While the current surgical robots use calibrated sensors to measure the position of the arms and tools, in this work we present a purely optical approach backed by artificial neural networks that provides a measurement of the tool position in relation to the patient's tissues.

Previous studies have evaluated the use of deep learning in the segmentation of medical data \cite{allan20192017}, however this research focuses on a particular conditional generative adversarial network (cGAN) called Pix2Pix and its potential use in field. Our novel contribution in this project was to build a Pix2Pix cGAN to recognize robotic arms in a surgical setting as a basis for an injury prevention system in robotic telesurgery.

\section{Theory}

\begin{figure}[!h]
    \centering
    \includegraphics[width= \linewidth]{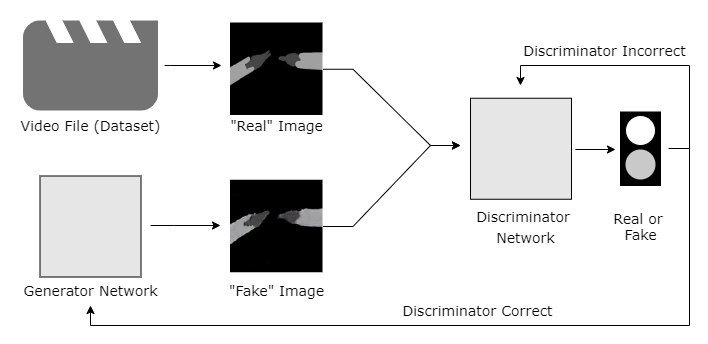}\centering
    \caption{High-level representation of cGAN algorithm}
    \label{fig:ganmodel}
\end{figure}

Conditional GANs combine a generative network that produces images from stochastic noise distributions, with a discriminator network that performs image recognition classification task \cite{yi2019generative, kazeminia2018gans, goodfellow2014generative}. These two networks compete against each other to see which network can become more accurate as shown in figure \ref{fig:ganmodel}. The generator starts off by creating random noise images and feeding them into the discriminator, along with sample images from the original data set. The discriminator then decides whether the image it’s fed is a ‘real’ picture - meaning from the data set - or a ‘fake’ image that was produced by the generator.

\begin{figure}[ht]
\begin{subfigure}{.45\textwidth}
    \includegraphics[width =\linewidth]{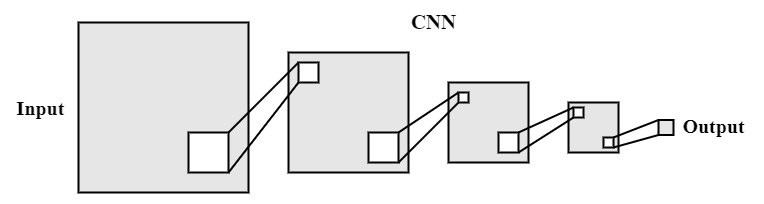} 
    \caption{Typical Convolutional Neural Network Architecture used for image processing, classification, and segmentation.}
    \label{fig:convnet}
\end{subfigure}
\begin{subfigure}{.45\textwidth}
  \centering
    \includegraphics[width = \linewidth]{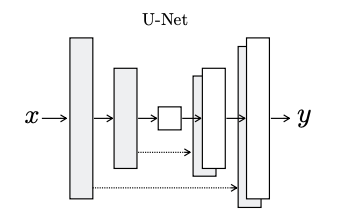} 
    \caption{U-Net encoder-decoder with skip connections between mirrored layers in the encoder and decoder stacks.}
    \label{fig:unet}
\end{subfigure}
\caption{Sub-networks operating within cGAN model}
\label{fig:nets}
\end{figure}

In both cases,  the model can be expressed in terms of minimizing a loss function \cite{mirza2014conditional}.
\begin{equation}
\begin{split}
\min_G \max_D V(D, G) = \mathbb{E}_{x \sim p_{data}(x)} [\log D(x \vert y)] + \\
\mathbb{E}_{z \sim p_z(z)} [\log(1 - D(G(z \vert y)))]
\end{split}
\label{ganeq}
\end{equation}
The discriminator is a convolutional neural network architecture (CNN) that breaks down images in order to learn how to recognize and identify important parts of the image such as edges, corners, and colors through a process illustrated in figure \ref{fig:convnet}. Convolutional Neural networks (CNN) are a subset of deep learning algorithms built to model the basic structure of the human primary visual cortex \cite{Lecun98gradient-basedlearning, jin2017deep}. In the case of image processing, they take images as inputs, learn features increasing in abstraction throughout the network layers, and then learn how these features relate to the specific image domain. If the network is fed an image from the generator, and it decides that the image is a fake, the generator takes that feedback and adjusts its weights in order to produce a more realistic image. The generator is a U-Net architecture that builds a dense embedding of an image using convolutional layers and expands that embedding into a new generated image \cite{kamrul2019u}. This process is shown in figure \ref{fig:unet}. Eventually, the generator that started off producing random images, begins producing images that seem real enough to fool the discriminator. If the discriminator is wrong in its conclusion (for example guessing that a fake image was real), it will adjust its own weights to better its accuracy for the future. 

\section{Data and Methods}

\subsection{Dataset}

The 2015 Endovis Challenge dataset used as the training data included videos of a mock gastrointestinal surgery in the form of 3 videos, each 44 seconds long \cite{grand}. The first video is endoscopic video footage of a robotic arm simulating a surgery in an ex-vivo setup. Each frame of the video has a corresponding hand-drawn label for the positioning of the right arm and left arm which makes up the other two video files (one video for the left arm segmentation, one for the right).

\subsection{Data Preparation}

This research utilized a PyTorch implemented Pix2Pix model written by Jun-Yan Zhu, Taesung Park, and Tongzhou Wang \cite{Isola_2017_CVPR}. PyTorch is a Python-based deep learning framework modeled after the Torch framework that uses multidimensional arrays as tensors. Pix2Pix is a Conditional GAN that is specifically used for image to image translation and segmentation. It takes images and their labeled segmentations and learns how to convert from one to another. Because the entire research was conducted using Colab (Google’s online Jupyter notebook), we were able to clone the Github repository to a Google drive and access the model from there.

\begin{figure}[!h]\centering
    \begin{subfigure}[t]{0.4\textwidth}
        \centering
        \includegraphics[width=.9\linewidth]{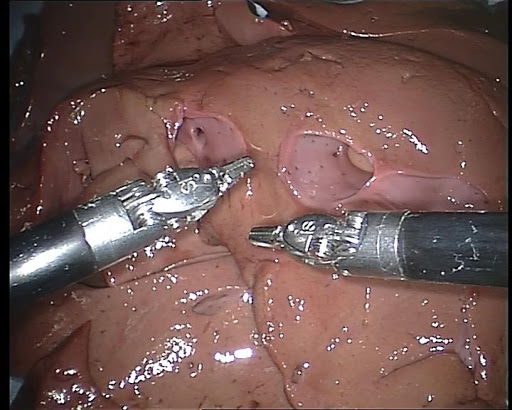}
        \caption{Sample image of endoscopic video feed}
        \label{subfig:videoframe}
    \end{subfigure}
    \qquad
    \begin{subfigure}[t]{0.4\textwidth}
        \centering
        \includegraphics[width=.9\linewidth]{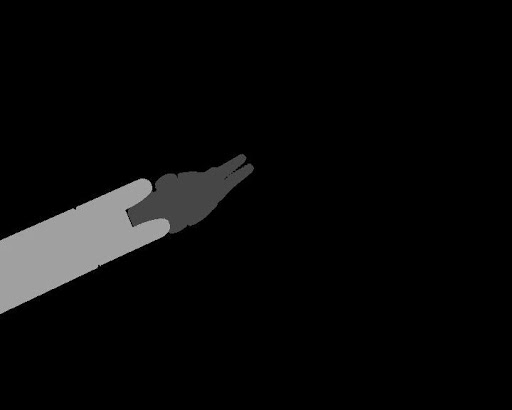}
        \caption{Left arm hand-drawn segmentation label for 4a}
        \label{subfig:leftseg}
    \end{subfigure}
    \qquad
    \begin{subfigure}[t]{0.4\textwidth}
        \centering
        \includegraphics[width=.9\linewidth]{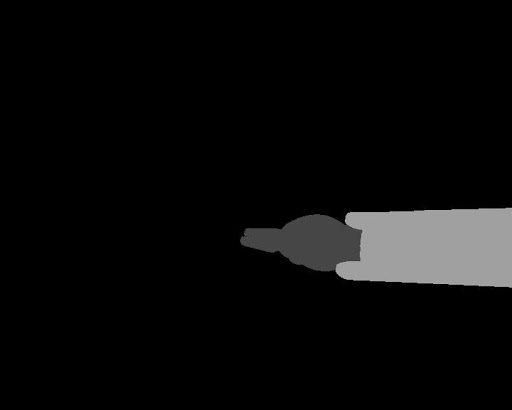}
        \caption{Right arm hand-drawn segmentation label for 4a}
        \label{subfig:rightseg}
    \end{subfigure}
    \begin{subfigure}[t]{0.4\textwidth}
        \centering
        \includegraphics[width=.9\linewidth]{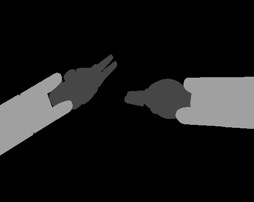}
        \caption{Combined segmentation label}
        \label{subfig:comboseg}
    \end{subfigure}
    
    \caption{Sample frames from EndoVis Challenge dataset}
\end{figure}

The model required the input data (images and labels) to be entered as singular paired images. We first split the video files into individual image frames pictured in figure \ref{subfig:videoframe}. Since the segmented label videos were separated by arm as shown in \ref{subfig:leftseg} and \ref{subfig:rightseg}, we combined them into one image with both arm segmentations show in \ref{subfig:comboseg}. The endoscopic pictures and combined segmentation labels were then stitched together to create an image with both frames side by side, which was then uploaded into the Google drive. This process was repeated 1107 times, for each frame of the video. This model was trained for 200 epochs, and was tested for accuracy every 5 epochs.

\section{Testing and Results}

\subsection{Generated Labels}
The main objective of this research was to produce labels for robotic surgery images. The primary results are the labels that were generated which are portrayed in table \ref{table:images}. The goal for the model was to create images that looked as similar as possible to the hand-drawn labels that were acquired from the dataset. For the input image in the first epoch of training, the respective label produced by the generator was noticeably blurry and there is a stark difference when compared to its hand-drawn label for that same image shown. By the 200th epoch, the model got even more accurate and the label produced was nearly identical to the hand-drawn label. Additional results can be found in Appendix \ref{appendix:a}.

\begin{table*}[h]
    \centering
    
    \begin{tabular*}{\textwidth}{|M{.13\textwidth}|P{.4\textwidth}|P{.4\textwidth}|}
          \hline
          \textbf{Description} & \textbf{Epoch 1} & \textbf{Epoch 200} \\
          
          \hline
          Endoscopic camera frame & \raisebox{-.5\height}{\includegraphics[width=.5\linewidth]{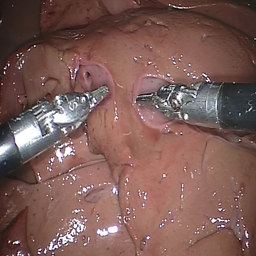}}  & \raisebox{-.5\height}{\includegraphics[width=.5\linewidth]{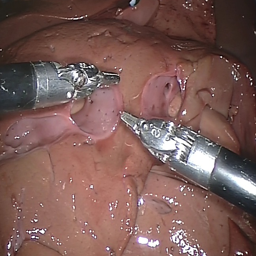}} \\
          
          \hline
          Hand drawn label & \raisebox{-.5\height}{\includegraphics[width=.5\linewidth]{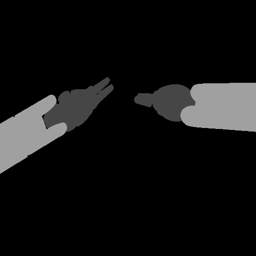}}  & \raisebox{-.5\height}{\includegraphics[width=.5\linewidth]{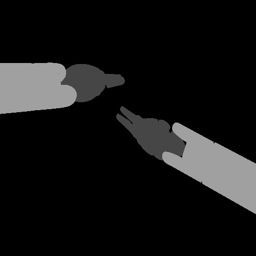}}  \\
          
          \hline
          Label created by generator & \raisebox{-.5\height}{\includegraphics[width=.5\linewidth]{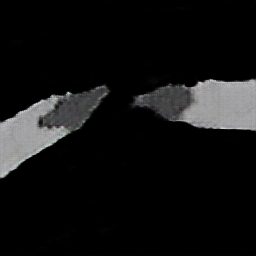}} & \raisebox{-.5\height}{\includegraphics[width=.5\linewidth]{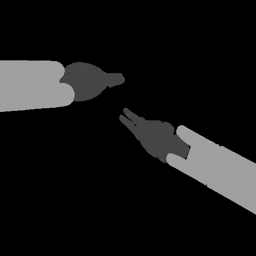}}\\
          
          \hline
          Subtracted difference between generated image and hand drawn image & \raisebox{-.5\height}{\includegraphics[width=.5\linewidth]{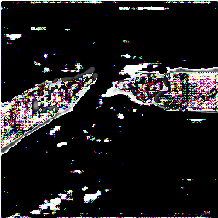}}  & \raisebox{-.5\height}{\includegraphics[width=.5\linewidth]{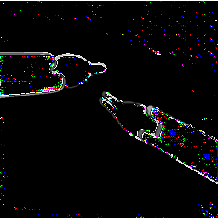}} \\
          
          \hline
          Histogram of per-pixel error &
          \raisebox{-.5\height}{\includegraphics[width=\linewidth]{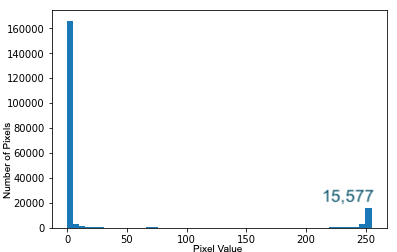}}  &
          \raisebox{-.5\height}{\includegraphics[width=       \linewidth]{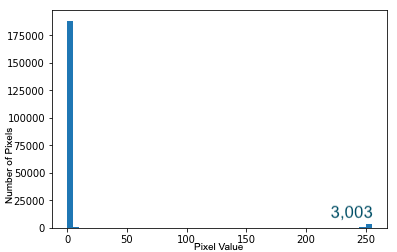}} \\
    
          \hline
          
    \end{tabular*}
    
    \caption{Comparative results of untrained (epoch 1) vs trained (epoch 200) model}
    \label{table:images}
\end{table*}

\clearpage


\subsection{Image Subtraction}
\begin{figure}[h]
    \centering
    \includegraphics[width = \linewidth]{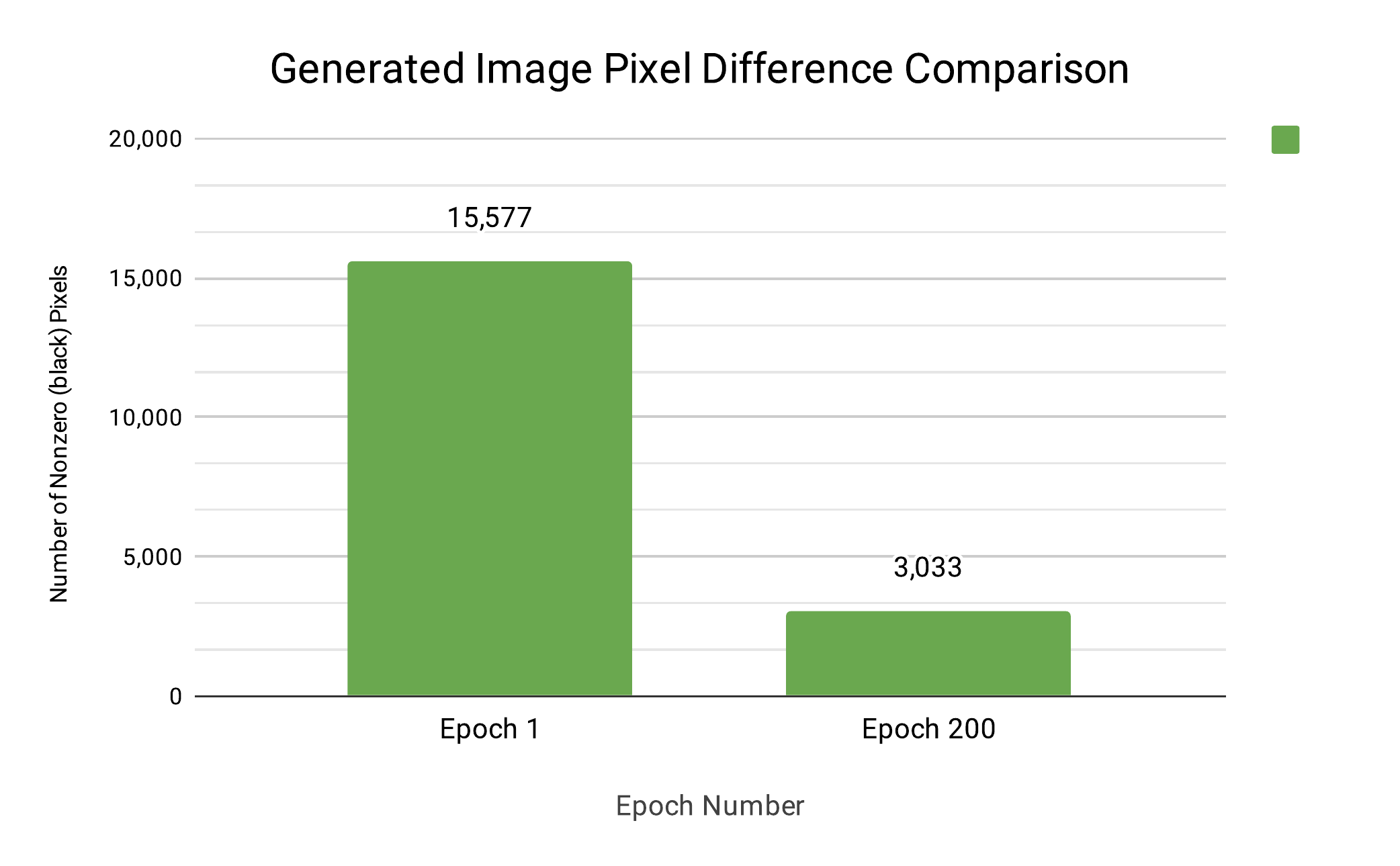}
    \caption{Direct comparison between non-zero pixel values from the untrained (epoch 1) and trained (epoch 200) subtracted difference images}
    \label{fig:pixel_diff}
\end{figure}

In addition to visually comparing the generated labels, we computed the mathematical differences between the trained model (epoch 200) and the untrained model (epoch 1). By subtracting the generated label from the hand-drawn label, we found the pixel differences between the two images. In table 1, the untrained model subtracted difference has a visibly large amount of noise in the image. While there is some noise on the background of the image (likely due to cropping) the majority of the pixel difference is centered in the robotic arms. This compared to the trained model in which the little noise that exists is mostly around the edges of the robotic arms. This data was then converted into a histogram in which we could see the pixel’s color concentration within the subtracted images. For the untrained data, the spike of white pixels represented by the '250' values, is noticeably higher than that of the trained data. This difference is highlighted in the per-pixel comparison located in figure \ref{fig:pixel_diff}. The numerical representation of the data is evidence for the advancement of the model and its sophistication in the matter of recognizing robotic arms, as we would expect the difference per pixel would decrease as images increased in similarity.\\

\section{Discussion and Future Research}
The network was able to perform very well on the surgical images with two arms, achieving near-perfect accuracy with the generator by epoch 200. The difference in the number of non-zero pixels shows a five-fold increase in accuracy. This supports the hypothesis that a Conditional Generative Adversarial Network has the capability to learn and reproduce what a surgical robotic arm looks like in a surgical setting. 

With the ability to segment and track the robotic arms, the next important piece of this research is the time factor. If the trained model took too long to process the images that it was given, then the entire premise of using it as a solution to the issue of latency in remote surgery would fail. To test this, we wrote a script that timed how long the model took to segment a single input image which ended up being 299 milliseconds. This time period is underneath the mark at which latency critically effects surgery, and thus affirms the applicability of this model.

\begin{figure}[h!]\centering
    \begin{subfigure}[h]{0.25\textwidth}
        \centering
        \includegraphics[width=.78\linewidth]{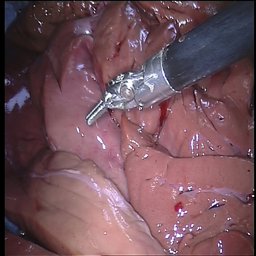}
        \caption{Single Arm Frame (raw footage)}
    \end{subfigure}
    \qquad
    \begin{minipage}[h]{.9\linewidth}
    \begin{subfigure}[h]{.43\linewidth}
        \centering
        \includegraphics[width=\linewidth]{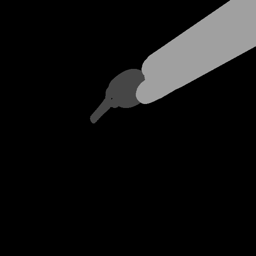}
        \caption{Single Arm Segmentation}
    \end{subfigure}
    \qquad
    \begin{subfigure}[h]{.43\linewidth}
        \centering
        \includegraphics[width=\linewidth]{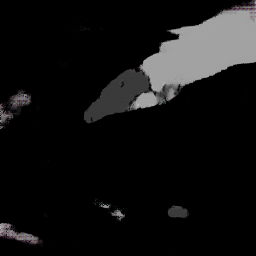}
        \caption{Single Arm Generated Segmentation}
    \end{subfigure}
    
    \caption{Test frames to check for adaptability of model}
    \label{fig:singleArm}
    \end{minipage}
\end{figure}

The findings presented in this research indicate that a neural network of the conditional generative adversarial architecture can be effectively used to teach a system how to recognize its own robotics limbs; however, there are some limitations of the model that need to be addressed before this system can be applied to a real surgery.

The dataset that we used for our model training was limited to images of two robotic arms moving around in gastrointestinal surgery. Because of this, the model learned that there were always going to be two arms in every image, and when it was tested on an image with only one robotic arm, the generator got confused and produced an inaccurate image (figure \ref{fig:singleArm}). 

The versatility that deep learning offers makes it possible to expand the training data to include images of a single-armed robot, and the model will in response learn how to recognize them. In fact, many issues regarding the scope of this project can be solved by adding to the training data and familiarizing the model with all types of robotic surgeries. For example, if a different attachment such as a stapler or forcep needs to be added, the model can be trained to recognize all of the necessary component by simply adding the respective images to the training data.

This network has the potential to allow the application of telesurgery both in places where high-speed fiber-optic connections are not available where latency is prevalent (such as underdeveloped countries, on a submarine, or in outer space) and in any places where lag and network connections are a risk factor. Telesurgery can be applied to the case where a wounded soldier needs a surgery that requires they are usually flown to the closest hospital, but for many war zones, these doctors are hard to find or too far away for the injured to reach. This will give medical professionals further reach to help patients, and can allow telesurgery to save lives in the years to come.

The aim of this research was to produce a system that could learn how to recognize robotic limbs, however, the potential of cGANs in surgery has a much larger scope that has yet to be explored. Applications in organ labeling to improve accuracy and tracking of other surgical instruments can all be achieved with neural networks and machine learning. In this research, we were able to produce a neural network that has the ability to track robotic arms and tell their position in a surgical context. By devising a system to detect when the arms are moving towards dangerous positions within patients, this research will provide the base for future research in applying telesurgery to places where high-speed fiber-optic connections are not available. 

This study was limited to video data of dual arm robotic surgeries, however future work would include a larger sample of data to ensure usability on a larger range of surgical procedures. Additionally, the two dimensional nature of the training images lacks sense of depth perception that may be necessary in an actual implementation of the remote surgical setup. Future projects can include further cross-validation of the model to ensure accuracy on diverse data sets as well as ensure an adequate sense of depth perception. The following steps would be to perform telesurgery simulations with programmed latency using the model overlay on the da Vinci instrument, and eventually clinical studies to really test how the model would interact in real time. 

Pre-trained models will be available open source for further research. 
All code can be found at \href{https://github.com/NeilSachdeva/RoboGAN}{Github Link}. Data set can be found at \href{https://endovissub-instrument.grand-challenge.org/Data/}{EndoVis Challenge data set link}.

\section{Declarations}
Conflict of Interest: Neil Sachdeva, William Hahn, Misha Klopukh and Rachel St. Clair declare that they have no conflict of interest
Consent Statements: This research study was conducted retrospectively from data obtained for public use.

\nocite{*}
{\small
\bibliographystyle{ieeetr}
\bibliography{bio.bib}
}

\section{Appendix}
\appendix{
\section{Additional results from model tests}
\label{appendix:a}

Camera Image \hfill Real Label \hfill Generated Label
\begin{figure}[!hb]
\setlength{\parindent}{0pt}
   \begin{minipage}{0.3205\linewidth}
     \setlength{\parindent}{0pt}
     \centering
     \includegraphics[width=\linewidth]{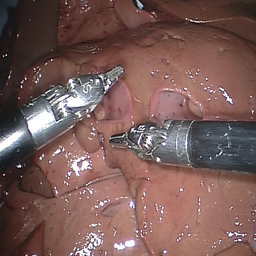}
   \end{minipage}
   \begin{minipage}{0.3205\linewidth}
     \setlength{\parindent}{0pt}
     \centering
     \includegraphics[width=\linewidth]{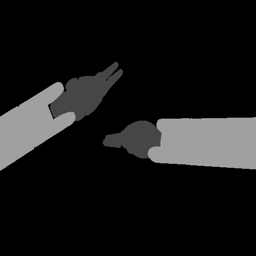}
   \end{minipage}
   \begin{minipage}{0.3205\linewidth}
     \setlength{\parindent}{0pt}
     \centering
     \includegraphics[width=\linewidth]{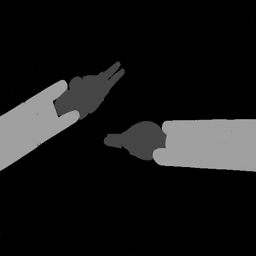}
   \end{minipage}
\end{figure}

\begin{figure}[!hb]
\setlength{\parindent}{0pt}
   \begin{minipage}{0.3205\linewidth}
     \setlength{\parindent}{0pt}
     \centering
     \includegraphics[width=\linewidth]{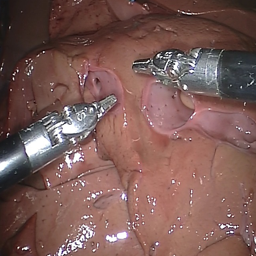}
   \end{minipage}
   \begin{minipage}{0.3205\linewidth}
     \setlength{\parindent}{0pt}
     \centering
     \includegraphics[width=\linewidth]{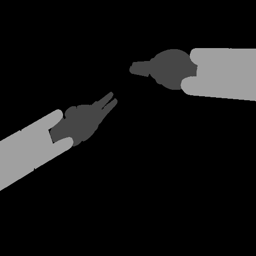}
   \end{minipage}
   \begin{minipage}{0.3205\linewidth}
     \setlength{\parindent}{0pt}
     \centering
     \includegraphics[width=\linewidth]{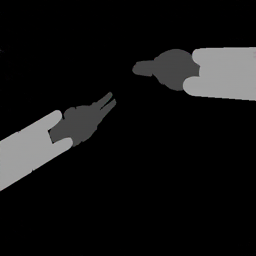}
   \end{minipage}
\end{figure}
}

\end{document}